\documentclass{article}

\PassOptionsToPackage{numbers, sort&compress}{natbib}
    \usepackage[preprint]{neurips_2023}

\usepackage[utf8]{inputenc} % allow utf-8 input
\usepackage[T1]{fontenc}    % use 8-bit T1 fonts
\usepackage{hyperref}       % hyperlinks
\usepackage{url}            % simple URL typesetting
\usepackage{booktabs}       % professional-quality tables
\usepackage{amsfonts}       % blackboard math symbols
\usepackage{nicefrac}       % compact symbols for 1/2, etc.
\usepackage{microtype}      % microtypography
\usepackage{xcolor}         % colors

\usepackage{pgf}
\usepackage{amsmath}
\usepackage{amssymb}

\usepackage{soul}
\usepackage{wrapfig}
\usepackage{multirow}
\usepackage{makecell}

\usepackage{algorithm}
\usepackage{algpseudocode}
\usepackage[color=yellow]{todonotes}

\usepackage{caption}
\usepackage[export]{adjustbox}   % <---
\usepackage{booktabs, tabularx} 

\newcommand{\weightlifter}{\includegraphics[width=0.016\paperwidth]{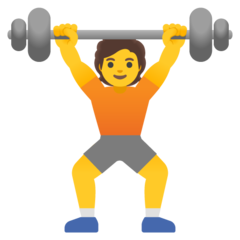}}

\renewcommand{\t}{\mathrm{t}}
\newcommand{\s}{\mathrm{s}}
\renewcommand{\a}{\mathrm{a}}
\renewcommand{\tt}{\tilde{\mathrm{t}}}
\newcommand{\ts}{\tilde{\mathrm{s}}}
\newcommand{\tx}{\tilde{x}}
\renewcommand{\L}{\mathcal{L}}
\newcommand{\D}{\mathcal{D}}

\title{HARD: Hard Augmentations for Robust Distillation}
\author{{Arne F. Nix\textsuperscript{1-2,*},
Max F. Burg\textsuperscript{2-3},
Fabian H. Sinz\textsuperscript{1-2, **}}
\\ \textsuperscript{1} Institute for Bioinformatics and Medical Informatics, University of Tübingen
\\ \textsuperscript{2} Institute for Computer Science and Campus Institute Data Science, University of Göttingen 
\\ \textsuperscript{3} Institute for Theoretical Physics, University of Tübingen
\\\\ \textsuperscript{*}\texttt{arne.nix@uni-goettingen.de},     \textsuperscript{**}\texttt{sinz@cs.uni-goettingen.de}
}

\begin{document}

\maketitle 

\begin{abstract}
    Knowledge distillation (KD) is a simple and successful method to transfer knowledge from a teacher to a student model solely based on functional activity.
    However, current KD has a few shortcomings: it has recently been shown that this method is unsuitable to transfer simple inductive biases like shift equivariance, struggles to transfer out of domain generalization, and optimization time is magnitudes longer compared to default non-KD model training.
    To improve these aspects of KD, we propose Hard Augmentations for Robust Distillation (HARD), a generally applicable data augmentation framework, that generates synthetic data points for which the teacher and the student disagree.
    We show in a simple toy example that our augmentation framework solves the problem of transferring simple equivariances with KD.
    We then apply our framework in real-world tasks for a variety of augmentation models, ranging from simple spatial transformations to unconstrained image manipulations with a pretrained variational autoencoder.
    We find that our learned augmentations significantly improve KD performance on in-domain and out-of-domain evaluation. 
    Moreover, our method outperforms even state-of-the-art data augmentations and since the augmented training inputs can be visualized, they offer a qualitative insight into the properties that are transferred from the teacher to the student.
    Thus HARD represents a generally applicable, dynamically optimized data augmentation technique tailored to improve the generalization and convergence speed of models trained with KD.\footnote{Code available at \href{https://github.com/sinzlab/HARD}{https://github.com/sinzlab/HARD}}
\end{abstract}

\section{Introduction}

Knowledge distillation (KD) methods \citep{Hinton2015,McClure2016, Zagoruyko} are powerful and flexible tools to transfer the knowledge of a given \emph{teacher} model to the transfer target, the \emph{student} model, without copying the weights. 
Instead, these methods match the student's functional activity (e.g. the softmax output) to that of the teacher for the presented inputs.
Hence, those methods are independent of architectural details and allow knowledge distillation to be applied in scenarios like model compression \citep{Bucil2006, Hinton2015}, continual learning \citep{Pan, Titsias2019, Benjamin2019}, or even neuroscience \citep{Li2019}, where traditional transfer learning would be impossible to use. 
KD methods also appear to be key to training new models that trade off inductive biases for more flexibility and more parameters \citep{Vaswani2017,Dosovitskiy2020, Tolstikhin2021} on smaller data \citep{Touvron2020,Chen2022,pmlr-v151-nix22a}.
However, \citet{pmlr-v151-nix22a} recently showed that current KD methods fail to transfer even simple equivariances between teacher and student. 
Additionally, previous work showed that KD leads to a larger gap between student and teacher on out-of-domain evaluation performance compared to within domain performance \citep{oquab2023dinov2,Beyer2021}, even in cases where the student almost perfectly matches the teacher \citep{Beyer2021} (see Table~\ref{tab:literature}).
This phenomenon is especially pronounced for particularly robust teachers \citep{oquab2023dinov2}.
Thus we expect that transferring robustness properties is a difficult problem for KD in general.

We hypothesize that KD methods are in principle capable of transferring most knowledge from a teacher to a student if the training data is chosen adequately.
We confirm this hypothesis on a small toy example (Section~\ref{sec:toy}), showing the importance of input data for KD.
Motivated by this demonstration, we propose our \emph{Hard Augmentations for Robust Distillation (HARD)} method, a general framework (Section~\ref{sec:framework}) to generate augmented training inputs which improve knowledge transfer by maximizing the distance between teacher and student while leaving the teacher's output unchanged.  
Consequently, our framework moves the input in directions that the teacher is invariant to but which are most challenging for the student.
Our experiments (Section~\ref{sec:exp}) show that our task-agnostic framework improves transfer effectiveness and thereby solves the problem of of KD not being able to transfer shift equivariance~\cite{pmlr-v151-nix22a}. 
Additionally, as part of our framework, we propose several parameterized augmentations (Section~\ref{sec:augmentors}) that can be integrated with most existing KD methods and are applicable to a variety of different computer vision tasks.
Finally, we demonstrate across multiple different models on the tasks of CIFAR10 and ImageNet that our framework learns interpretable augmentations that improve KD to the same level and in many cases even beyond established data augmentation methods, even when evaluated in an out-of-domain setting.
\section{Related Work}

\begin{figure}
    \centering
    \vspace{-0.2cm}
    \includegraphics[width=\linewidth,trim={4.cm 0cm 5cm 0},clip]{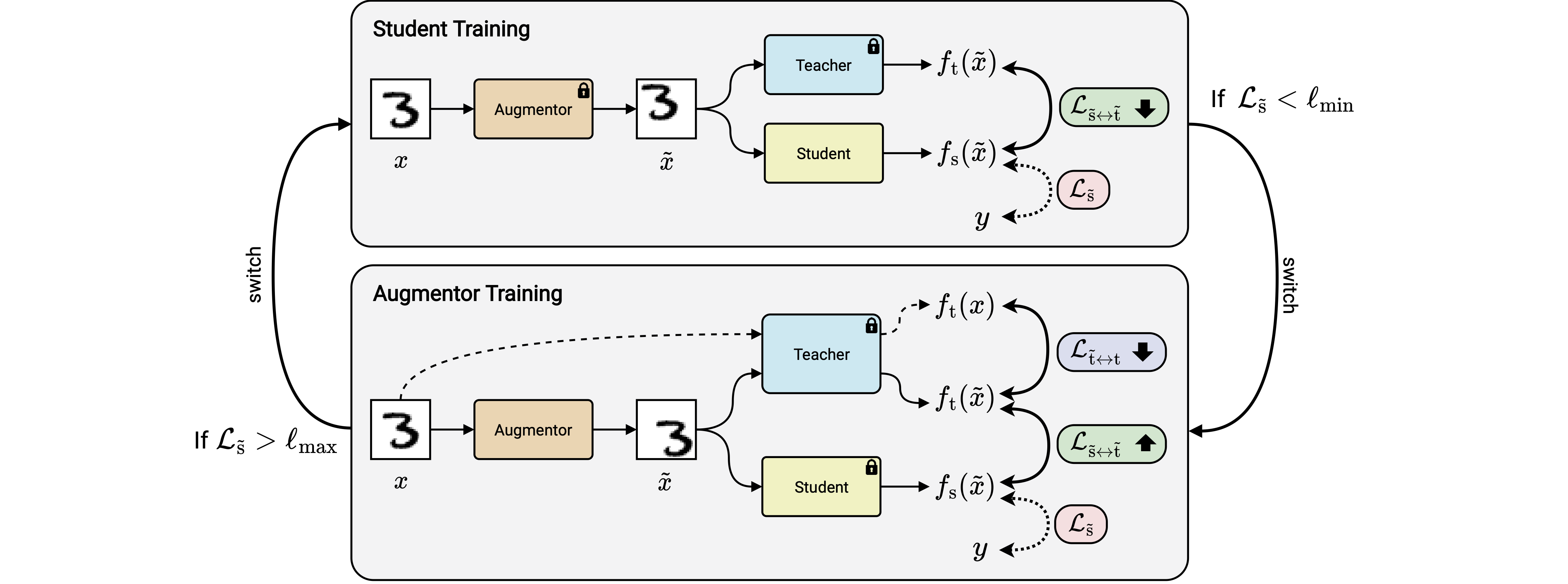}
    \caption{Our task-agnostic HARD framework switches between training the student to match the teacher and training the augmentor to generate new samples on which the student underperforms while maintaining high teacher performance. We optimize the augmentor and student in interchanging phases through a student-teacher loss $\L_{\ts \leftrightarrow \tt}$ and teacher-teacher loss $\L_{\tt \leftrightarrow \t}$. We switch between the two phases by comparing the default loss $\L_{\ts}$ on augmented data to pre-defined thresholds.}
    \label{fig:framework}
    \vspace{-0.2cm}
\end{figure}

There is a long tradition in using data augmentations to artificially extend training data for deep learning models and particularly in computer vision, be it through adding Gaussian noise, random crops, shifts, flips, or rotations \citep{NIPS2012_c399862d,engstrom2017rotation}.
In recent years, data augmentations became more complex \citep{hendrycks2019augmix,muller2021trivialaugment,cubuk2020randaugment,Zhang2017,yun2019cutmix,jackson2019style}, employing a multitude of different heuristics with the aim to improve generalization and in some cases also out-of-domain performance \citep{hendrycks2019augmix}. 
A particularly popular augmentation method is \emph{Mixup} \citep{Zhang2017}, which randomly interpolates two input samples and their labels respectively. 
Similarly, \emph{Cutmix} \citep{yun2019cutmix} combines two input images by pasting a random crop of one image on top of the other. 
Also,  many studies use parameterized augmentations optimized to improve a given objective \citep{Cubuk2018,Rusak2020,Hendrycks2021, Zietlow2022}, and some even optimize the augmentations to improve on an adversarial objective \citep{Volpi2018,Behpour2019, Zhang2019, Zhang2019a, Zhao2020,Gong2021, Antoniou2022}, however, without applying them for knowledge transfer.

In KD, applying data augmentations is a very effective tool to improve matching between student and teacher \citep{Wang2020, Beyer2021} and optimizing on a meta level can be useful to aide the teaching \citep{Pham2021}.
Similar to our work, \citet{Rashid2021, Haidar2022, Zhang2022} utilized adversarial objectives to optimize data augmentations for KD, however, they were solely focused on natural language processing tasks and do not optimize the augmentations towards invariance.

% On top of the impressive results of these data augmentation methods for generalization, there is also lots of research 
% ur method is inspired by the large body of work investigating different types of data augmentations: 
% \citet{Golan2020} optimize new data points that lie by construction at the decision boundary of two models, i.e. at a point at which the decisions of both models do not coincide anymore.
Inspired by this large body of work we formulate a task-agnostic framework containing only one building block that is specific to the data-domain -- the instantiation of the augmentor model generating the augmented data samples -- for which we offer a variety of reasonable model choices based on spatial transformer modules \citep{Jaderberg2015}, Mixup \citep{Zhang2017}, and variational autoencoders \citep{Kingma2014a,Child2020,lee2022autoregressive}. % and successfully apply our framework to image classification.

% Recently, many researchers showed that synthetic images generated by diffusion models~\cite{dm,ramesh_hierarchical_2022,saharia_photorealistic_2022,ldm} improve model accuracy in downstream classification tasks when used for data augmentation~\cite{roy_diffalign_2022,zhang_expanding_2022,he_is_2022,trabucco_effective_2023,shipard_diversity_2023,azizi_synthetic_2023,ghalebikesabi_differentially_2023, akrout_diffusion-based_2023}. However, as generating a single image with diffusion model requires multiple seconds, this model type is generally to slow to be applied for our proposed dynamic data augmentation technique.

\section{Input Data Matters for Functional Transfer}
\label{sec:toy}

\begin{figure}
    \centering
    \begin{tikzpicture}
    \draw (0, 0) node[inner sep=0] {
    \input{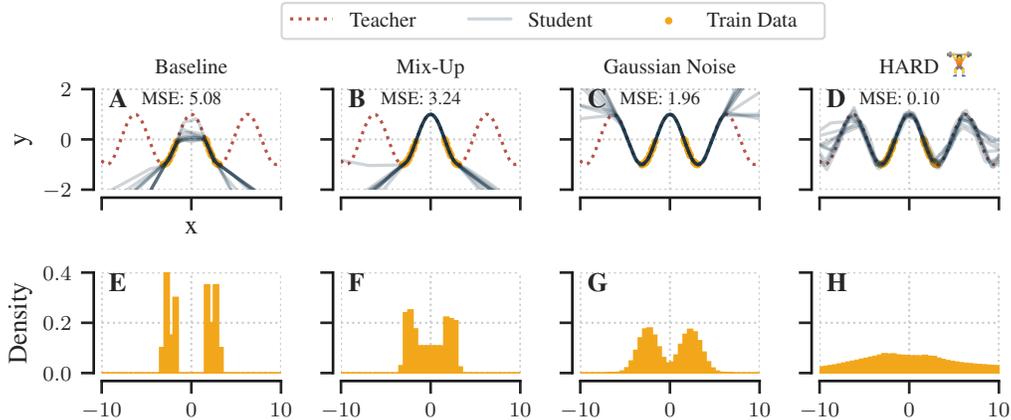}
    };
    \draw (5.9, 1.97) node { \footnotesize\weightlifter};
\end{tikzpicture}
    \vspace{-0.7cm}
    \caption{Fitting the student, a three-layer ReLU MLP, to the teacher function, $\cos(x)$, for $10,000$ iterations. We show results for 10 random seeds (\textbf{A}-\textbf{D}) and the distribution of (augmented) training inputs as a normalized histogram (\textbf{E}-\textbf{H}).
    We compare baseline (no augmentations) with Mixup, Gaussian noise and an HARD-optimized noise distribution. We report mean-squared-error (MSE) on 100 test inputs sampled from $\mathcal{U}_{[-10,10]}$.
    }
    \label{fig:sin}
    \vspace{-0.3cm}
\end{figure}

We hypothesize that the choice of input data is crucial to successfully knowledge distillation and we illustrate the impact of training data by a simple toy example.
To demonstrate this, consider a simple KD task in which we instantiate the teacher-model by the true function $f_{\t}(x) = \cos(x)$ and the student $f_{\s}(x)$ by a three layer Multilayer Perceptron (MLP) with ReLU activation \citep{agarap2018deep}.
We use input data $x$ 
% from $\mathcal{X}_\mathrm{train} \subset (-\pi, -\frac{\pi}{2}) \cup (\frac{\pi}{2},\pi)$, 
chosen such that it does not capture the teacher's $\cos(x)$ periodicity (orange points in Figure~\ref{fig:sin}A). 
Simple KD does neither interpolate between the given training points nor extrapolates beyond them (Figure~\ref{fig:sin}E).
Hence the student neural network does not learn the teacher's periodicity and fails to interpolate and extrapolate beyond the training data (Figure~\ref{fig:sin}A).

Augmenting the training data with more helpful inputs $\tilde{x}$ and teacher labels $f_{\t}(\tilde{x}) = \cos(\tilde{x})$ could mitigate this problem. 
One method successfully applied to KD \citep{Beyer2021} is to extend the input data through Mixup~\citep{Zhang2017}. 
When applying this to our illustrative example, we create new training inputs $\tilde{x}$ through linear interpolation between pairs of input points $\tilde{x}=(1-\alpha) x_1 + \alpha x_2$ (Figure~\ref{fig:sin}F), 
 %,\ 0 \le \alpha \le 1$, 
and recording the corresponding teacher responses $f_{\t}(\tilde{x}) = \cos(\tilde{x})$.
Thus, the student learns to interpolate between training points, but mixup does not enhance extrapolation (Figure~\ref{fig:sin}B).

To generate datapoints that would interpolate and extrapolate beyond already available training points, we could simply augment by adding Gaussian noise $\epsilon$ to the available data points, $\tilde{x}= x + \epsilon$, %,\ \epsilon \sim \mathcal{N}(\mu=0,\sigma=1)$.
hence interpolating and extrapolating beyond the training data (Figure~\ref{fig:sin}G). This strategy helps our student to match the teacher also outside the original training regime (Figure~\ref{fig:sin}C).
However, the student only improves within a fixed margin that is determined by the noise distribution's mean and variance.

We could obviously improve interpolation and extrapolation by increasing the noise distribution's variance or shifting it's mean, however, as we move to a high dimensional image input space ($x \in \mathbb{R} \rightarrow \Vec{x} \in \mathbb{R}^N$) it becomes unclear how to heuristically select helpful new samples and at the same time random exploration strategies become computationally infeasible.
Instead, we propose to optimize a parameterized augmentation to efficiently generate new, hard training samples on which the student lacks performance, as here the student could improve the most. In our toy example, we illustrate this by optimizing the Gaussian's parameters (mean and variance) according to our augmentation framework \emph{HARD}, which we will present in the next section. 
This provides us with a noise distribution which we use to draw new helpful training examples $\tx$ that transfer inter- and extrapolation to the student network (Figure~\ref{fig:sin}D,H). Overall, this toy example shows that learning hard augmentations to select new helpful data points is crucial to efficiently improve extrapolation beyond the training distribution.
% TODO

% While lifting this constraint by heuristically or randomly exploring different $\mu$  and $\sigma$ seems to be an easy solution in our 1D example, it is  unclear how to select these in the general case of high-dimensional inputs with a search-space that cannot be tractably explored at random.
% Instead, we propose to optimize a parameterized augmentation (in this 1-D example we could for instance learn $\mu$ and $\sigma$) to generate synthetic datapoints such that we can maximize the learning benefit for our student, matching 
% the teacher on a much larger area .
% , however, it is unclear how to select helpful datapoints, especially once we move to high dimensions where the input search space quickly becomes infeasibly large ($x \in \mathbb{R} \rightarrow \Vec{x} \in \mathbb{R}^N$).

\section{Learning Hard Augmentations for Robust Distillation (HARD)}

\label{sec:framework}

Our task-agnostic HARD framework learns augmenting training images to most efficiently help knowledge-transfer from a teacher to a student model. Our method requires three main components: a teacher model with frozen parameters, a student model that should learn knowledge from the teacher, and a parameterized augmentation model that learns to augment images such that most of the teacher's knowledge is transferred to the student.

% I think we should mention somewhere how we deal with catastrophic forgetting. Maybe also worth adding how we initialize / warmup, explaining that we start from the default KD case? I also wonder if this section misses some important details or would we add them to the appendix?

    % \paragraph{Teacher-student loss}
    In classical KD methods\citep{Hinton2015}, the objective is to minimize a distance $\D\left[f_{\s}(x),f_{\t}(x)\right]$ between the student's activation $f_\s(x)$ and the teacher's activation $f_\t(x)$ on given inputs $x \in \mathbb{R}^n$.
    Usually, this would be the Kullback-Leibler divergence between the softmax distributions of teacher and student.
    Unfortunately, only considering  training data could miss properties of the teacher (eg. shift invariance) that might be crucial for generalization (see Section~\ref{sec:toy} for an illustrative example). 
    To resolve this issue, we learn a parametrized augmentation model $g_\a$ to generate new input data points $\tx = g_\a(x)$ transferring such invariance properties from the teacher to the student.
    Hence, we define a \textit{teacher-student loss} considering the more general case of matching student and teacher on augmented inputs $\tx\in \mathbb{R}^n$:
    \begin{align}
        \L_{\ts\leftrightarrow \tt} = \D\left[f_{\s}(\tx), f_{\t}(\tx)\right]\ . 
    \end{align}

    % The student is optimized as usual to minimize this objective.
    % However, the augmentor $g_\a$ is trained to maximize the same objective that the student is trying to minimize.
    % This leads us to an adversarial framework \citep{Goodfellow2014} for augmentation.    
    
    % \paragraph{Teacher-teacher loss}
    To specifically transfer the teacher's invariance properties to the student, we propose a \textit{teacher-teacher loss} pushing the augmentor towards generating data points on which the teacher is invariant, 
    % We achieve by minimizing the distance between the teacher's output for the original data $x$ and its output for the augmented data $\tilde{x}$:
    \begin{align}
        \L_{\tt \leftrightarrow \t} = \D\left[f_{\t}(\tx), f_{\t}(x)\right]\ ,
    \end{align}
    as these are often useful augmentations for generalization.
    % \paragraph{The algorithm} 
    Using both of these losses, we optimize the augmentor's parameters $\theta_\a$ to generate augmented samples on which the teacher results in similar activations but the student differs from them (Figure~\ref{fig:framework} top) and simultaneously we optimize the student's parameters $\theta_\s$ to perform well on those augmentations (Figure~\ref{fig:framework} bottom):
    \begin{align}
        \max_{\theta_\a} \  \lambda_\s \L_{\ts \leftrightarrow \tt} - \lambda_\t \L_{\tt \leftrightarrow \t} 
        \text{\quad\quad and \quad\quad} \min_{\theta_\s} \  \L_{\ts \leftrightarrow \tt}\ .
    \end{align}
    Here, $\lambda_\s$ and $\lambda_\t$  trade off the loss terms and are treated as hyper-parameters. % thatand were set by an empirical grid-search.
    % The optimization of student and augmentor can either be done jointly or separately, which is possible because $\L_{\tt \leftrightarrow \t}$ prevents a diverging solution for our augmentor.
    We train both components separately switching from training the augmentor to training the student when the student's performance on augmented data gets worse than a pre-defined threshold ($\L_{\ts} > \ell_{\max}$) and we switch back from student to augmentor training when the student's performance on augmented data surpasses a pre-defined threshold ($\L_{\ts} < \ell_{\min}$; Figure~\ref{fig:framework}).
    To prevent catastrophic forgetting, we save augmentors at every switch and employ an augmentor randomly chosen out of the set of previously saved augmentors in each iteration when training the student.

\subsection{The augmentor models}
\label{sec:augmentors}
    To generate new input data points it is important to choose an augmentor that suits the desired application and is powerful enough to generate useful augmentations.
    Usually, we do not know a priori what useful augmentations are and thus should try to allow as much flexibility as possible.
    Additionally, some variance over augmentations could benefit the transfer.
    Thus, all augmentors in our study introduce randomness in the model by adding Gaussian noise into the computation of the augmentation through the reparametrization trick \citep{Kingma2014a}.
    While our framework is universally applicable across domains, choosing an effective augmentation model likely needs to be addressed for each task individually.
    In our experiments, we use the following augmentor models:

\begin{figure}
    \centering
    \vspace{-0.5cm}
    \includegraphics[width=0.97\textwidth]{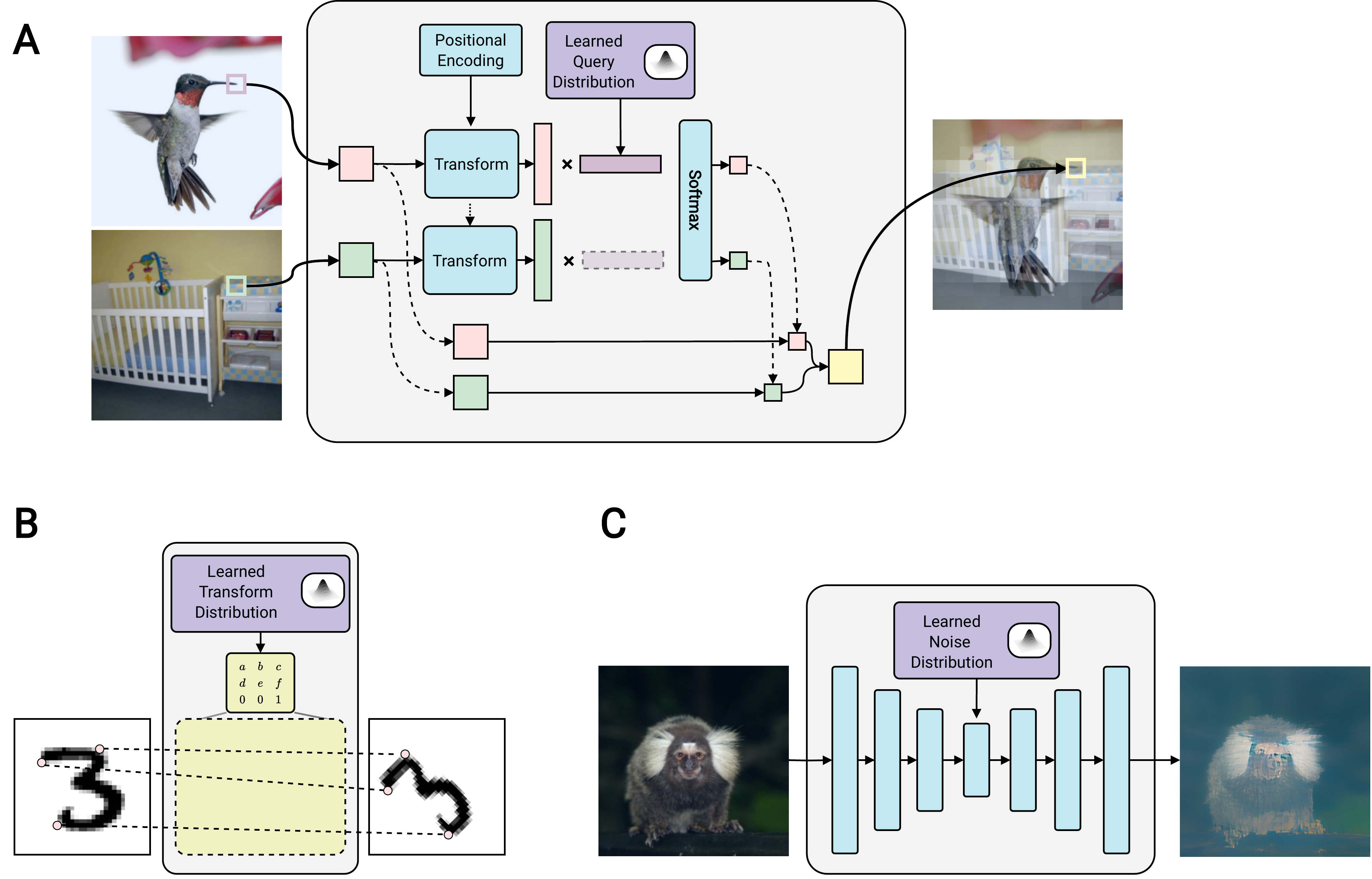}
    \caption{Illustration of the augmentor models used in our experiments. 
    \textbf{(A)} HARD-Mix: Image-dependent patch-wise interpolation of multiple images. 
    \textbf{(B)} HARD-Affine: Learned distribution of affine transformations in the pixel coordinates. 
    % \textbf{(C)} HARD-Style: Learned distribution of style embeddings, applied to an image through instance-normalization in a pretrained network. 
    \textbf{(C)} HARD-VAE: Finetuning (parts of) a pretrained VAE.}
    \label{fig:augmentors}
\end{figure}

\paragraph{HARD-Affine}
    In the simplest model, we limit the augmentations to affine transformations of the coordinate grid of pixel locations, i.e. shifts, rotations, scalings, and shears of images. 
    Models implementing such transformations are known as \emph{spatial transformers} \citep{Jaderberg2015}.
    % Spatial transformers are parameterized modules for manipulations of an input image. 
    We leverage this model for our augmentor by learning a distribution over the entries of an affine transformation matrix $\vartheta \in \mathbb{R}^{2\times 3}$ that defines the transformation of the sampling grid, i.e. a transformation that maps the pixel positions from the original image to the augmented image (Figure~\ref{fig:augmentors}A).
    % This augmentor has only 42 learnable parameters (one fully-connected layer).
    % , but allows for  augmentations such as cropping, translation, rotation, scale, and skew.
    
\paragraph{HARD-Mix}
    Additionally we consider a slightly more complex augmentor model, which is an adaptive variant of the commonly used Mixup \citep{Zhang2017} and Cutmix \citep{yun2019cutmix} augmentations. 
    However, instead of randomly sampling the ratio and cutout position that are used to combine images, we learn how to combine the images dependent on the input images.
    We achieve this by performing a patch-wise projection of the input image, followed by comparing each patch with the same query vector sampled from a learned distribution (Figure~\ref{fig:augmentors}B). 
     We normalize similarities for each patch over each group of images and use the resulting weights to combine the original image patches, giving a combined image.
    This mechanism allows our augmentor to decide which features of which image are shown to the student, enabling it to explore the interpolated space between images systematically, instead of randomly.
     As it would not make sense for the teacher to be invariant to an interpolation as it is generated by HARD-Mix, we do not consider the teacher-teacher-loss $\L_{\tt \leftrightarrow \t}$ in this case and optimize student and augmentor jointly instead.
    
% \paragraph{HARD-Style}
%     This augmentation method manipulates texture, contrast and color while preserving the shape and semantic information of the image content. 
%     To achieve this we adapt the method of \emph{random style augmentation}~\cite{jackson2019style}.
%     Following their method, we manipulate the style of a given image by passing it through a pretrained image-to-image network with frozen weights \citep{ghiasi2017exploring} and by injecting a style embedding through conditional instance normalization \citep{dumoulin2017a}.
%     \citet{jackson2019style} sample the style embedding from a fixed Gaussian distribution with mean and covariance estimated from a set of embedded drawings. 
%     Instead, our method learns the mean and diagonal covariance of this distribution (Figure~\ref{fig:augmentors}C).
%     This allows the augmentation to optimize and reflect styles that are hard for the student as training progresses.

\paragraph{HARD-VAE}
    To lift constraints further, we wanted to use a more powerful augmentor that could generate a large variety of images across the entire image-space. 
    As the augmentor has to generate new samples on-the-fly during the student training, the generation process needs to be very fast, limiting the choice of useful generative models.
    % To impose even less restrictions and allow augmentations that could explore a large approximation of image space, we propose the use of a generative image model as the augmentor. 
    % Potentially any generative model that allows for image-conditioned generation could be used for this purposed.
   % However, it is crucial that the generation process is very fast, since it needs to be applied on-the-fly during the student training phase.
   For this reason, we focus on variants of the variational autoencoder architecture \citep{Kingma2014a}, allowing for good image reconstructions which can be achieved reasonably fast in a single forward pass (Figure~\ref{fig:augmentors}D).
   For CIFAR, we choose the \emph{very deep VAE} \citep{Child2020} model, which we finetune by solely optimizing parameters of the posterior network from layer 10 onward in the decoder.
    For the experiments on ImageNet, we use a Residual-Quantized VAE~(RQ-VAE)~\citep{lee2022autoregressive} pretrained on ImageNet, which we finetune in its entirety and add a noise vector on the latent state.
    Hence, as training progresses, the model changes from generating plain reconstructions of a given image to input conditioned generations that serve as our augmentations.

\section{Experiments}
\label{sec:exp}
\begin{table}[]
    \caption{MNIST (columns ``Centered'') and MNIST-C (columns ``Shifted'') test accuracies (mean and standard error of the mean across 4 random seeds) comparing KD without augmentation and our HARD-Affine method to Orbit transfer \citep{pmlr-v151-nix22a}, which also learns and transfers equivariances. The left two columns show the transfer results from a small CNN teacher to a MLP student. The right columns show analogous experiments between a ResNet18 teacher and a small ViT student. The best performing transfer is shown in bold for each column. Examples of our HARD-Affine learned data augmentations shown on the right. 
    We include the controls \textit{Random Affine} and \textit{MNIST-C Shifts} (marked by italics).
    % ``HARD-Affine (\textit{Shifted})'' shows best-performing settings for the shifted test set (i.e. not selected based on centered Validation).
    }
    \vspace{0.3cm}
    \begin{tabularx}{\linewidth}{@{}c X @{}}
        \begin{tabular}{lcccc}
    \toprule
      & \multicolumn{2}{c}{CNN $\rightarrow$ MLP}  & \multicolumn{2}{c}{ResNet18 $\rightarrow$ ViT}  \\
      \cmidrule(lr){2-3}
      \cmidrule(lr){4-5}
      Method &  Centered & Shifted &  Centered & Shifted  \\
      \midrule
      Teacher only &  99.0 \footnotesize $\pm$ 0.0 & 91.3 \footnotesize $\pm$ 0.5 & 99.5 \footnotesize $\pm$ 0.0 & 92.8 \footnotesize $\pm$ 0.5 \\
      Student only  &  98.4 \footnotesize $\pm$ 0.0 & 35.2 \footnotesize $\pm$ 0.7 & 98.3 \footnotesize $\pm$ 0.0 & 40.4 \footnotesize $\pm$ 0.8 \\
      \ \ \textit{+ Random Affine } &  \textit{92.1 \footnotesize $\pm$ 0.6} & \textit{81.0 \footnotesize $\pm$ 2.0} & \textit{95.4 \footnotesize $\pm$ 0.3} & \textit{90.4 \footnotesize $\pm$ 1.0} \\
      \ \ \textit{+ MNIST-C Shifts}  &  \textit{98.1 \footnotesize $\pm$ 0.1} & \textit{86.5 \footnotesize $\pm$ 0.3} & \textit{98.5 \footnotesize $\pm$ 0.0} & \textit{93.7 \footnotesize $\pm$ 0.2} \\
      \midrule
      Orbit \citep{pmlr-v151-nix22a} & \textbf{98.8} & \textbf{95.2} &  98.4 & \textbf{84.0}\\
      KD  &  98.6 \footnotesize $\pm$ 0.0 & 40.3 \footnotesize $\pm$ 0.6 & 98.6 \footnotesize $\pm$ 0.1 & 44.7 \footnotesize $\pm$ 1.9\\
      \midrule
        \ \  + HARD\weightlifter-Affine &  98.6 \footnotesize $\pm$ 0.1 & 68.9 \footnotesize $\pm$ 2.5 & \textbf{99.2 }\footnotesize $\pm$ 0.0 & \textbf{84.1} \footnotesize $\pm$ 2.3\\
        % \ \   \phantom{+ HARD \weightlifter } \textit{(Shifted)} &  \textit{98.5 \footnotesize $\pm$ 0.1} & \textit{91.7 \footnotesize $\pm$ 0.4} & \textit{99.2 \footnotesize $\pm$ 0.0} & \textit{89.3 \footnotesize $\pm$ 0.8}\\
      \addlinespace
      \bottomrule
    \end{tabular}
        & \hspace{0.5cm} \includegraphics[width=0.5\linewidth, valign=c,trim={0.cm 0.5cm 0cm 0cm},clip ]{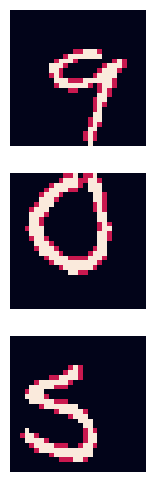}
    \end{tabularx}
    \vspace{-0.2cm}
    \label{tab:mnist}
\end{table}

\subsection{Transferring equivariance}
% \paragraph{Transferring Shift Equivariance}
\label{sec:MNIST}
    For our initial experiment, we reproduce the setup from \citet{pmlr-v151-nix22a} to test whether we can transfer the inductive bias from a shift equivariant teacher, CNN and ResNet18 \citep{He2015}, to a student that does not have this inductive bias built into its architecture: a Multi-Layer Perceptron (MLP) and a Vision Transformer~(ViT)~\citep{Dosovitskiy2020}.
    When training the students and teachers by themselves on standard MNIST \citep{deng2012mnist} training data, we observe a small drop of generalization performance (-0.6\% and -1.2\%) between teacher and student on the MNIST test set and a large gap (-56.1\% and -52.4\%) when we evaluate on a version of the test set in which digits were randomly shifted \citep{Mu2019}.
    As another baseline, we applied plain KD to transfer shift equivariance from teacher to student. 
    Consistent with the findings of \citet{pmlr-v151-nix22a}, we only observe a small improvement on the centered (+0.2\% and +0.3\%) and the shifted (+5.1\% and +4.3\%) test sets, which likely result from the centered training data we use for transfer. 
    
    We then test if combining KD with our augmentations produced by HARD-Affine would outperform these baselines.
    The resulting student model improves significantly on shifted inputs (+28.6\% and +39.4\%) compared to plain KD and the generated images clearly show that the augmentor learns to shift the digits within the image.
    % To verify the potential performance ceiling of our method, we additionally report results for hyper-parameter settings (different learning-rate) that were chosen based on performance on the shifted test data instead of the centered validation set.
    % This shows us that we could potentially get very close to the teacher's performance on this metric.
    Compared to \citet{pmlr-v151-nix22a} our approach outperforms their results on the ViT task but, while improving the out-of-domain generalization by 28.6\% over baseline, stays behind the Orbit performance on the MLP task.  
    % and gets close to their result with an MLP student when we use our best hyper-parameter configuration. 
    This demonstrates that our method while acting on fewer parts of the network compared to Orbit 
    % (ours only acts on the input of the student) 
    and while being a more general method,
    % (can be added to most existing functional transfer methods)
    can improve or reach better performance when it comes to transferring invariances, and can be generalized to bigger datasets, as we show below. 

    We verify that the student's performance improvement is specifically due to our data generation framework in two control experiments. 
    The first experiment (Random Affine) augments the training inputs of a stand-alone student model with a random affine transformation akin to our augmentor model, but using transformation parameters sampled uniformly from a pre-defined, reasonably constrained range (i.e. ensuring the digit is always fully visible).
    This student performs well on the shifted test set, however, performance significantly degrades on the centered test set. 
    In comparison, our HARD-Affine model is unconstrained  and learns  more useful augmentations, leading to better performance on the centered test sets.
    
    In our second control (Shifts) we asked how much data augmentation could improve the performance in the best case (without KD).
    For this, we augment the inputs by the same random shifts that were applied to obtain the shifted test data, leading to great improvements on the shifted test set. 
    However, our learned augmentations achieve scores in a similar range on the shifted evaluation and outperform its results on the centered test set.

\subsection{Transfer on natural images}
\label{sec:Natural}
After demonstrating that our method successfully captures the difference between teacher and student and bridges a gap in inductive bias, we now want to test whether this effect holds up in more realistic scenarios. 

\paragraph{CIFAR experiments}
\label{sec:CIFAR}
\begin{table}[th]
    \caption{Test accuracies on the CIFAR10 test set. Standard error of the mean is reported where available across three different seeds. Best transfer is highlighted in bold. The ResNet101$^\ast$ models were pretrained on ImageNet. Examples of augmented test images from ResNet18$\rightarrow$ViT experiments with samples across different iterations are shown to the right.}
    
    \begin{minipage}[t]{0.64\linewidth}
    \begin{adjustbox}{valign=b, width=\linewidth}
     \begin{tabular}{lccc}
    \toprule
    & \makecell{ResNet18  \\ $\downarrow$ \\ ViT} & \makecell{ResNet101$^\ast$ \\ $\downarrow$ \\ ViT }  & \makecell{ResNet101$^\ast$  \\ $\downarrow$ \\ ResNet18} \\ % & \makecell{WRN-40-5 \\ $\downarrow$ \\ WRN-16-1} \\
    \midrule
    Teacher only & 92.5 \footnotesize $\pm$ 0.0 & 95.5 & 95.5 \\ %& 92.6 \\
    Student only & 68.5 \footnotesize $\pm$ 0.5 & 68.5 \footnotesize $\pm$ 0.5 & 78.5 \\ %& 68.4 \\
     \ \ + Standard Aug. & 78.3 \footnotesize $\pm$ 0.4 & 78.3 \footnotesize $\pm$ 0.4  & 92.6 \\ %& 85.8 \\
     \ \ + Random Affine Aug. & 58.9 \footnotesize $\pm$ 0.4 & 58.9 \footnotesize $\pm$ 0.4 & 79.3 \\ %& 62.2 \\
     \midrule
    KD & 67.9 \footnotesize $\pm$ 0.1 & 68.5 & 84.4 \\ %& 70.3 \\
    \ \ + Standard Aug. & 80.9 \footnotesize $\pm$  0.1 & 79.3 & 93.3 \\ %& \textbf{87.4}\\
     \midrule
    \ \ + HARD\weightlifter-Affine & \textbf{87.8 \footnotesize $\pm$ 0.8} & 84.4 & 93.5 \\ %& 86.2 \\
    \ \ + HARD\weightlifter-VAE & 81.9 \footnotesize $\pm$ 0.4 & 81.2 & 91.0 \\ %& 79.3 \\
    \ \ + HARD\weightlifter-VAE-Affine & \textbf{87.6 \footnotesize $\pm$ 0.6}  & \textbf{87.1} & \textbf{94.0}  \\ %& 85.2 \\
      \bottomrule
    \vspace{.05ex}
    \end{tabular}
    \end{adjustbox}
    \end{minipage}
    \begin{minipage}[t]{0.35\linewidth}
    \includegraphics[width=\linewidth, trim={0.9cm 2.5cm 0.8cm 2.0cm},clip]{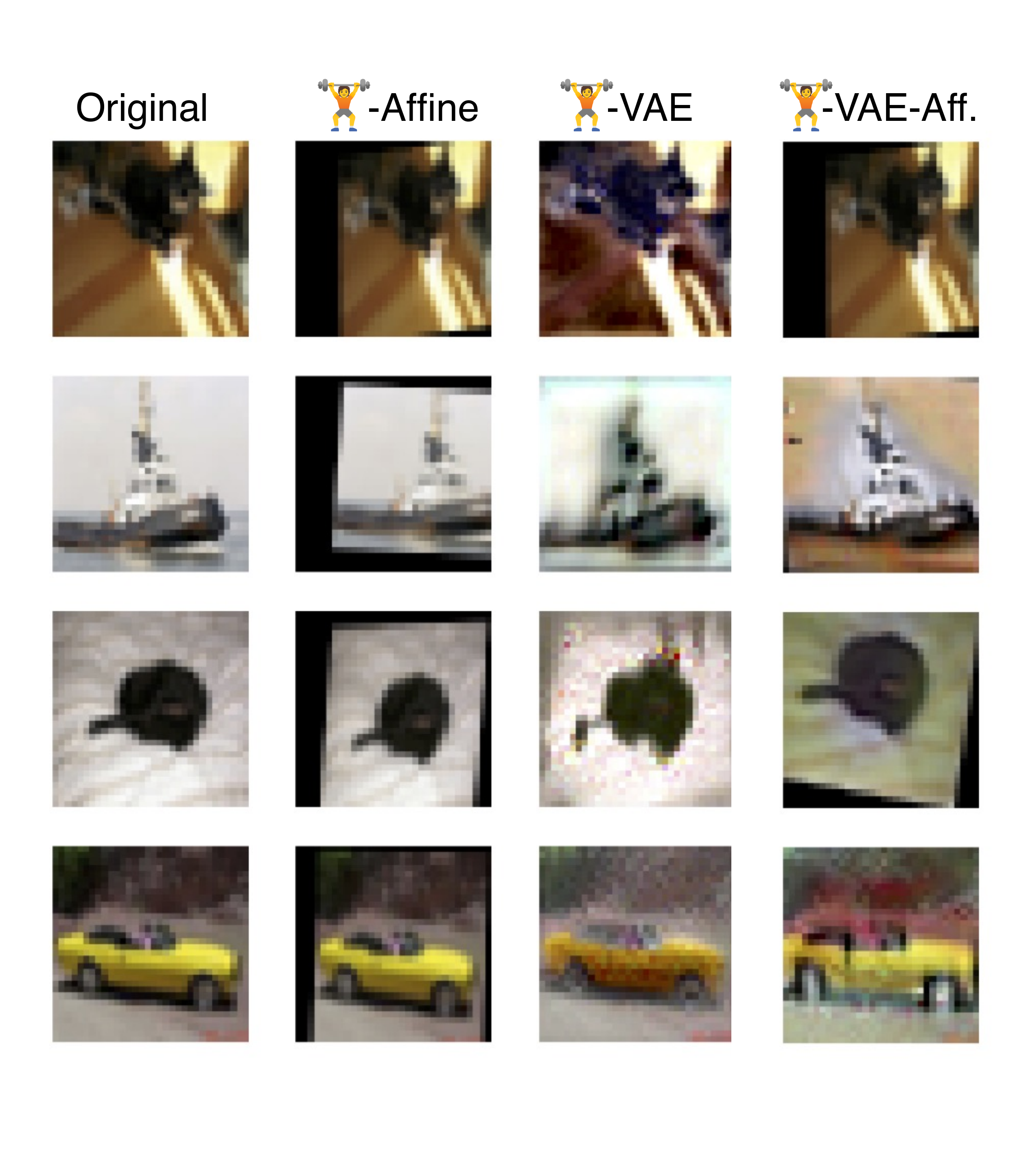}
    \end{minipage}
    \label{tab:cifar}
\end{table}

We begin by applying our framework to CIFAR10 \citep{Krizhevsky} on three different KD scenarios (see Table~\ref{tab:cifar}). 
Specifically, we test scenarios where the student lacks an inductive bias (ResNet18$\rightarrow$ViT), where the teacher has more capacity and access to data than the student (ResNet101$^\ast$ $\rightarrow$ResNet18), and to scenarios combining both properties (ResNet101$^\ast$ $\rightarrow$ViT).
For all experiments, we keep the experimental setup as close to our previous MNIST experiments as possible (see Appendix~\ref{app:setup} for details).

We start by establishing baselines by training only the teacher and only the student models on the data and evaluating default KD. We observe that on this small data set a small ResNet18 performs better (78.5\% accuracy) than a larger ViT (68.5\%), likely because of the ResNet's superior inductive bias on this task and small data set. Next, we find that adding default data augmentations (random rotations, cropping, horizontal flips) to the student baselines significantly boosts performance to 92.6\% and 78.3\% for the ResNet18 and ViT, respectively. Adding these default augmentations to typical KD leads to a great performance boost, too (see Table \ref{tab:cifar}).

Given that adding default data augmentation to KD already leads to a substantial performance boost, it is particularly noteworthy that the data augmentations learned by HARD-Affine outperform this baseline for the ViT.
Qualitatively, the augmented images exhibit a large variety of spatial transformations, suggesting that a difference in these examples lead to the observed performance boost (Table~\ref{tab:cifar}, right). % in performance on these examples is what separates student and teacher the most.

We then investigated performance of our HARD-VAE augmentation strategy and found performance improvement over the KD + standard augmentations baseline for transfer to the ViT (+1.0\% and +1.9\%) student.
However, inspecting the augmented images indicates that our augmentor lacks the expected shifts of object positions, but rather learns stylistic changes in the image (Table~\ref{tab:cifar}, right). This motivated us to combine HARD-Affine and HARD-VAE augmentation resulting in best performance (up to +7.8\%) for all teacher-student pairings (HARD-VAE-Affine in Table~\ref{tab:cifar}) and the resulting images demonstrate variability in both style and spatial alignment (Table~\ref{tab:cifar}, right).

\paragraph{ImageNet experiments}

Having established our methods' performance for CIFAR10, we extend our results to classification on ImageNet \citep{deng2009imagenet}.
Here we aim to distill a ResNet50 \citep{He2015} teacher, trained with Deep-augment and AugMix data augmentations \citep{Hendrycks2021}, into a smaller ResNet18 and ViT-S (small vision transformer variant) \citep{Dosovitskiy2020} that we want to be particularly robust to natural image corruptions,. 
% Here we focus on two settings:
The distillation into ResNet18 allows us to investigate the capability for model compression, because ResNet18 is a smaller network compared to ResNet50, but with a similar architecture.
Distillation into a ViT-S architecture with a patch-size of 14 tests additionally if KD transfers the ResNet50's inductive bias of shift equivariance on a larger dataset.

\begin{table}[th]
\centering
\begin{minipage}[c]{0.7\textwidth}
\adjustbox{width=\textwidth}{
\begin{tabular}{lcccccc}
     \toprule
\multicolumn{1}{c}{\multirow{2}{*}{}} & \multicolumn{3}{c}{ResNet50 $\rightarrow$ ResNet18} & \multicolumn{3}{c}{ResNet50 $\rightarrow$ ViT-S} \\ 
      \cmidrule(lr){2-4}
      \cmidrule(lr){5-7} 
\multicolumn{1}{c}{} & Val & ReaL & V2 & Val & ReaL & V2 \\ 
     \midrule
Teacher & 75.8 & 83.1 & 63.7 & 75.8 & 83.1 & 63.7 \\
Student & 70.7 & 78.1 & 57.4 & 73.2 & 79.4 & 60.3 \\
     \midrule
KD & 70.7 & 78.7 & 58.1 & 75.3 & 82.8 & 62.9 \\
     \midrule
\ \ + HARD\weightlifter-Affine & \textbf{71.6}& \textbf{79.5} & 58.6 & 74.9 & 82.3 & 62.4 \\
\ \ + HARD\weightlifter-Mix & 71.4 & 79.4 & 58.6 & 75.7 & 83.0 & 63.3 \\
 \ \ + HARD\weightlifter-VAE & 71.0 & 78.9 & \textbf{58.7} & \textbf{75.8} & \textbf{83.1} & \textbf{63.5} \\ 
\bottomrule
\end{tabular}
}
  \end{minipage}\hfill
  \begin{minipage}[c]{0.27\textwidth}
    \caption{In-domain evaluation for ImageNet: reporting Top-1 accuracy in \% on ImageNet-Validaton \citep{deng2009imagenet}, ImageNet-ReaL \citep{beyer2020we} and ImageNet-V2 \citep{recht2019imagenet} with KD from a robust ResNet50 \citep{Hendrycks2021} teacher to ResNet18 (columns 2-4) and ViT-S (columns 5-7) students.}
    \label{tab:indomain}
  \end{minipage}
\end{table}

\begin{table}[th]
\centering
\caption{In-domain evaluation for ImageNet: Reporting Top-1 accuracy in \% on ImageNet-A \citep{hendrycks2021nae}, ImageNet-R \citep{Hendrycks2021}, ImageNet-Sketch \citep{wang2019learning} and ImageNet-Style \citep{Geirhos2018} and mean-corruption-error on ImageNet-C (lower is better) \citep{Hendrycks2019a}.}
\label{tab:outdomain}
     \adjustbox{width=\textwidth}{
\begin{tabular}{lcccccccccc}
\toprule
\multicolumn{1}{c}{\multirow{2}{*}{}} & \multicolumn{5}{c}{ResNet50 $\rightarrow$ ResNet18} & \multicolumn{5}{c}{ResNet50 $\rightarrow$ ViT-S} \\ 
      \cmidrule(lr){2-6} 
      \cmidrule(lr){7-11} 
\multicolumn{1}{c}{} & Im-A & Im-R & Im-C $\downarrow$ & Sketch & Style & Im-A & Im-R & Im-C $\downarrow$ & Sketch & Style \\ 
     \midrule
Teacher & 3.8 & 46.8 & 53.0 & 32.6 & 21.2 & 3.8 & 46.8 & 53.0 & 32.6 & 21.2 \\
Student & 1.6 & 30.0 & 88.1 & 18.4 & 4.4 & \textbf{8.0} & 26.3 & 78.1 & 13.8 & 6.6 \\
     \midrule
KD & 1.6 & \textbf{40.2} & 69.2 & 26.0 & 13.4 & 3.3 & 45.0 & 56.8 & 29.6 & 18.7 \\
     \midrule
\ \ + HARD\weightlifter-Affine & 1.5 & 38.2 & 73.1 & 24.9 & 10.4 & 3.4 & 40.8 & 62.2 & 26.2 & 14.5 \\
\ \ + HARD\weightlifter-Mix & \textbf{1.8} & 39.9 & \textbf{68.8} & \textbf{26.1} & \textbf{13.7} & 3.5 & \textbf{45.4} & \textbf{56.2} & 29.9 & \textbf{19.2} \\
\ \ + HARD\weightlifter-VAE & 1.7 & 39.5 & 72.5 & 25.8 & 12.1 & 3.4 & \textbf{45.4} & 57.4 & \textbf{30.7} & 18.1 \\ 
\bottomrule
\end{tabular}
}
\end{table}

We evaluate on common test sets for both in-domain (ID) \citep{beyer2020we,recht2019imagenet} and out-of-domain (OOD) \citep{Hendrycks2019a,hendrycks2021nae,Hendrycks2021,Geirhos2018,wang2019learning} generalization performance (Tables~\ref{tab:indomain}~and~\ref{tab:outdomain}, respectively).
To properly investigate the extrapolation abilities of KD training, we trained a strong KD baseline by applying several data augmentations: 
we randomly switch between Cutmix~\citep{yun2019cutmix} and Mixup~\citep{Zhang2017}, each drawing their interpolation weight from a $\beta$-distribution with $\alpha=1$, as well as AugMix~\citep{hendrycks2019augmix} augmentations.
For the standalone student training, we additionally apply various lighter data augmentations (Cutmix with $\alpha=1$, Mixup with $\alpha=0.1$, and Trivialaugment~\citep{muller2021trivialaugment}). Since we ask how KD can be improved in a setting of limited resources, we run our experiments an order of magnitude shorter than proposed for the state-of-the-art in KD \citep{Beyer2021} (200 epochs for all ResNet18 and 150 epochs for all ViT-S experiments).  
For student and KD models, we perform a small grid search over learning-rate and weight-decay hyperparameters.
We then train the models with our HARD framework based on the hyperparameters of our best performing KD setting. 
The augmentor-specific settings are selected through a small grid-search in the ResNet18 setting (for details see Appendix~\ref{app:setup}). 

We first evaluate the ID performance of our methods (Table~\ref{tab:indomain}) beginning with the standalone teacher and student baselines, which reveal a larger performance gap between the ResNet18 student and the ResNet50 teacher compared to the ViT-S student (5.1\% and 2.6\% on the ImageNet validation set, respectively). 
Plain KD significantly reduces this gap for the ViT-S (+2.1\% performance improvement compared to standalone).
For the ResNet18 student KD achieves only small (0.7\% V2) improvements or no improvements (0.0\% Val), even though the initial gap between teacher and student is larger.
Applying HARD-Affine, HARD-Mix and HARD-VAE augmentation on this task improves over plain KD across most augmentation models and test sets with student performance gains of up to 0.9\% for ResNet18 (HARD-Affine) and 0.6\% for ViT-S (HARD-VAE). For ViT-S, our best-performing HARD-VAE method even matches the teacher's performance on 2 out of 3 test sets.

For the OOD setting (Table~\ref{tab:outdomain}), we observe that the initial gap between student and teacher is larger than on ID data across all data sets (up to 35.1\% difference), except for Im-A in the ViT-S setting. 
The aggressive data augmentations we apply for the plain KD baseline favor OOD performance, hence it is expected that plain KD results in good performance improvement over the standalone baseline (up to 21.3\% imporvement on Im-C).
All three HARD approaches transfer some of the teacher's generalization abilities leading to improvements on a number of students and data sets, however, HARD-Affine fails to reach the KD performance in both settings and HARD-VAE underperforms for the ResNet18 student in these OOD scenarios. However, HARD-Mix and HARD-VAE (for ViT-S) outperform plain KD on several test sets and are roughly on par on all others, across the board. Given that we chose a very strong baseline by applying aggressive state-of-the-art data augmentations we find these results especially encouraging.

\section{Discussion}

\begin{wrapfigure}[24]{r}{0.5\textwidth}
    \includegraphics[width=0.5\textwidth]{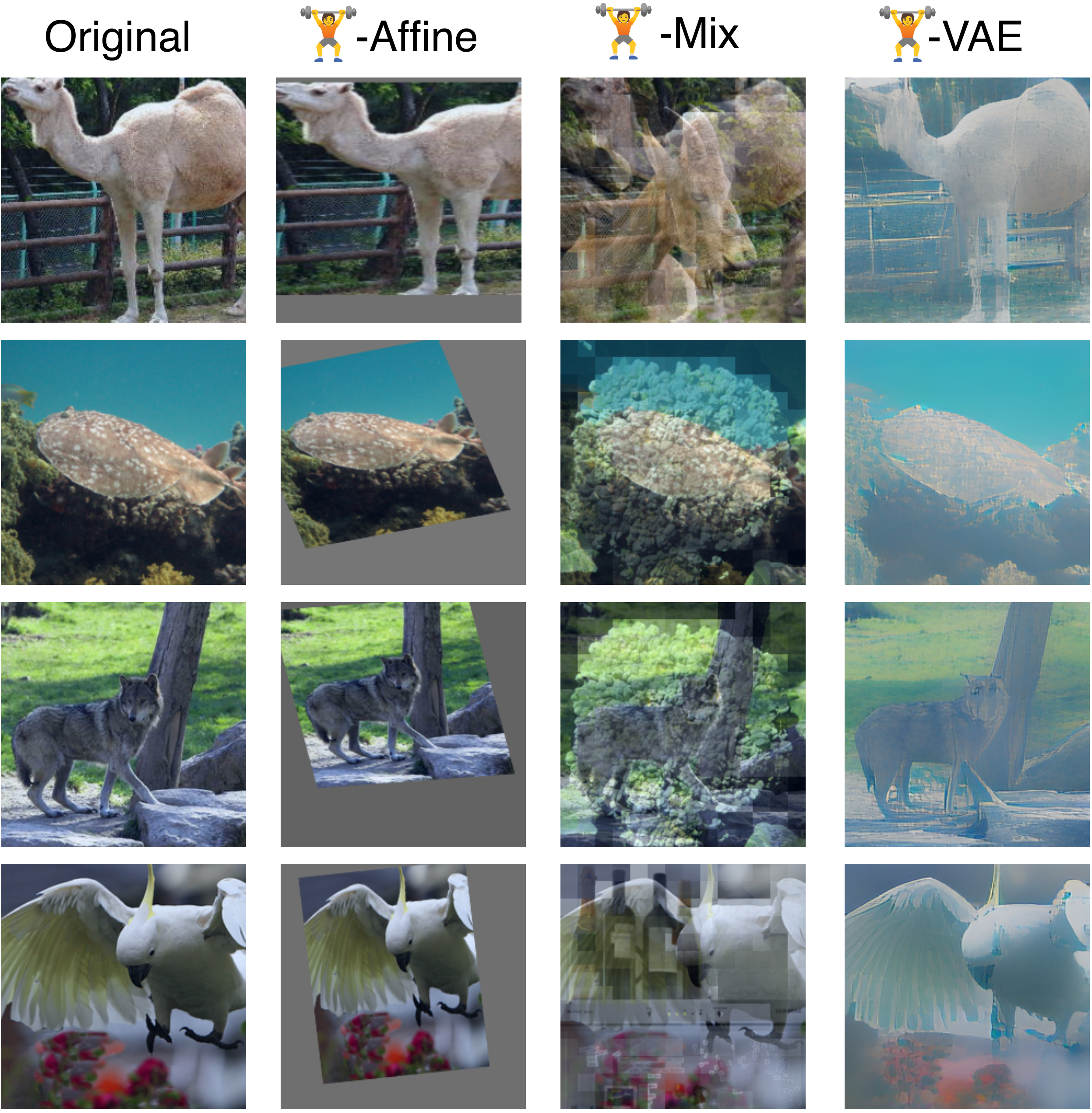}
    \caption{Example augmentations applied to images of the ImageNet validation set obtained from augmentor models in the ViT-S setting at the end of training.}
    \label{fig:imagenet_augs}
\end{wrapfigure}

\paragraph{Interpretability} HARD enables us to gain insight into the distillation mechanism as the augmented images illustrate the knowledge that is transferred (Figure~\ref{fig:imagenet_augs}).
% One of HARD's advantages is that it gives us an insight into the distillation mechanism, by showing us the knowledge being transferred through the augmented input examples. 
% Qualitatively analyzing the augmented images of the ImageNet experiments reveals some interesting properties of our augmentors.
As expected, HARD-Affine learns to downscale images to shift and rotate the images such that the object in the image is shown in different places (row 2-4 in Figure~\ref{fig:imagenet_augs}) and scales such that the images is cropped (row 1).
As HARD-Mix is a dynamically learnable extension of mixup, it either merges two objects into the same picture (row 1 and 4), especially if they are not in the same position, or uses one image to change the style (row 2) or background (row 3) of another. 
Finally, HARD-VAE mostly impacts the style of an image and additionally adds small distortions to specific image regions, which is noticable by the altered image brightness and the blurring of some high-frequency features. 

\paragraph{Limitations and broader impact}
State-of-the-art knowledge distillation typically deals with huge models (billions of parameters) and incredibly long training times (>9,000 epochs) \citep{Beyer2021,dehghani2023scaling}.
In comparison, our study is computationally lightweight in requiring approximately 400 A100GPU days across all our experiments. We believe exploring even more flexible augmentor models with a semantically meaningful latent space as for example diffusion models~\cite{dm,ramesh_hierarchical_2022,saharia_photorealistic_2022,ldm} could improve our proposed methods even further. However, generating a single image with out-of-the-box diffusion models requires multiple seconds. This is prohibitively long, so leave exploring their usability in our proposed dynamic data augmentation technique for future work.
In general, KD allows us to distill smaller models that perform similar to large foundation models. 
Improving the distillation process to be more efficient lowers the barrier of applying KD across labs with various compute budget and decreases environmental impact.
At the same time, transferring generalization abilities effectively and consistently results in smaller distilled models that are appealing to use, thus we would expect such smaller models to be used abundantly hence lowering the general carbon footprint for model usage.
In conclusion, our study proposes avenues to efficiently improve KD in terms of performance, efficiency, and hence environmental impact. 

\section{Conclusion}
In this work we introduced a general, task-agnostic, and modular framework to extend knowledge distillation by learnable data augmentations.
The augmentation models are optimized to generate inputs on which teacher and student disagree,  keeping the teacher's predictions unchanged at the same time. 
We show that these augmentations can solve the issue of KD and transfer equivariance properties, even in cases where the teacher's inductive biases are distinct from the student's.
We further demonstrate that our learned augmentations achieve performance competitive to established classical data augmentation techniques even when student and teacher share similar inductive biases. 
Overall our framework offers a powerful tool that enhances transfer performance and offers a unique insights into the transferred knowledge through its interpretable augmentations.

% \newpage

\subsubsection*{Acknowledgements}

% We thank all reviewers for their constructive and thoughtful feedback. 
Furthermore, we thank Felix Schlüter for his helpful insights into evaluation problems as well as Mohammad Bashiri, Pawel Pierzchlewicz and Suhas Shrinivasan for helpful comments and discussions.
The authors thank the International Max Planck Research School for Intelligent Systems (IMPRS-IS) for supporting Arne Nix and Max F. Burg.

This work was supported by the Cyber Valley Research Fund (CyVy-RF-2019-01), by the German Federal Ministry of Education and Research (BMBF) through the Tübingen AI Center (FKZ: 01IS18039A), by the Deutsche Forschungsgemeinschaft (DFG) in the SFB 1233, Robust Vision: Inference Principles and Neural Mechanisms (TP12), project number: 276693517, and funded by the Deutsche Forschungsgemeinschaft (DFG, German Research Foundation) – Project-ID 432680300 – SFB 1456.
FHS is supported by the Carl-Zeiss-Stiftung and acknowledges the support of the DFG Cluster of Excellence “Machine Learning – New Perspectives for Science”, EXC 2064/1, project number 390727645.

\bibliographystyle{plainnat}
\bibliography{main}

\newpage

\appendix

\section{Setup Details}
\label{app:setup}

Our experiments on MNIST were meant to reproduce \citet{pmlr-v151-nix22a} and thus follow their setup exactly, using the same training setup and model architectures. 

\subsection{CIFAR10 Experiments}
\paragraph{Training}
We train on the entire CIFAR10 dataset (excluding 10\% held-out as validation set) for 300 epochs with a batch-size of 256.
As an optimizer, we use Adam \citep{Kingma2014} with a learning rate of 0.0003 and an L2-regularization of 2$\cdot$10$^{-9}$. 
Our training begins with a linear warmup of the learning rate for 20 epochs.
The validation accuracy is monitored after every epoch and if it has not improved for 20 consecutive epochs, we decay the learning rate by a factor of 0.8 and restore the previously best performing model.
The training is stopped prematurely if we decay five times.

\paragraph{Models}
The different models we use generally follow the standard architecture and settings know from the literature.
For the ViT, we use a smaller variant of it on the CIFAR task. 
It consists of six layers and eight attention heads throughout the network.
The dropout rate is set to 0.1 and the hidden dimension is chosen as 512 in all places. 

\paragraph{KD and HARD}
After initial experiments on MNIST, we decided to use a softmax temperature of 5.0 for all experiments involving KD. 
We furthermore rely solely on the KL-Divergence loss to optimize our model.
For the experiments with our augmentation framework, we have the same settings as before for the student (KD) training and separate settings for the augmentor training.
There we have different settings depeding on whether we use the VAE augmentor (or the Affine augmentor).
There we reduce the batch-size to 160 (128) and a learning-rate of 0.0001 (0.05). 
We initialize both augmentors to perform an identity transformation, i.e. the VAE is taken pretrained from \citet{Child2020}.
The thresholds for switching are set as $\ell_{\min}=10\% (5\%)$ and $\ell_{\max}=60\% (40\%)$.
The train modi are switched if the threshold is surpassed for 5 consecutive iterations. 
Both $\lambda_\s$ and $\lambda_\t$ are set to 1 for the experiments. 
For the experiment ResNet101$^\ast$ $\rightarrow$ ResNet18, we found a slightly different setting to be more effective with $\ell_{\min}=$5\% and $\ell_{\max}= 40\%$ and a switch only happening if the threshold is surpassed for 10 consecutive iterations.

\subsection{ImageNet Experiments}
\paragraph{Baseline Training}
In general, all our ImageNet experiments follow a similar setup. 
We train with a batch-size of 512 samples using the Lion optimizer \citep{chen2023symbolic} with a linear learning-rate warmup to a defined initial learning-rate. 
Afterwards, we anneal the learning-rate following a cosine schedule \citep{loshchilov2017sgdr} with a final value of 0.
The training runs for 200 epochs for all ResNet18 experiments and 150 epochs for the ViT-S experiments.
Throughout the training, the validation accuracy is monitored on a heldout set consisting of samples randomly chosen from training set, making up 1\% of the total number of samples. 
The validation performance is used to pick the best performing epoch throughout training for final evaluation and the best hyperparameters during grid-search.
We train at a resolution of 224 pixels with random resizing and cropping, as well as random horizontal flips applied in all trainings.
All training runs are performed with automatic mixed precision and 8bit optimization \citep{dettmers2022optimizers}.

\paragraph{Student Training}
After a grid-search, we found that for the standalone student training, an optimization with learning-rate 0.0001 with weight decay 0.1 for the ResNet18 student and learning-rate 0.00005 with weight decay 0.001 for the ViT-S student worked best.  
For both students, we apply light augmentations during training with Mixup ($\alpha=0.2$) \citep{Zhang2017} and CutMix ($\alpha=1.0$) \citep{yun2019cutmix}.
For the ViT-S baseline, we additionally apply Trivial-Augment \citep{muller2021trivialaugment} and randomly erase pixels from the input image \citep{zhong2017random} with a probability of 0.1.
We optimize the standard cross-entropy loss with additional label-smoothing \citep{7780677} mixed in with a factor of 0.1.

\paragraph{KD and HARD}
As described in the main paper, the configuration for the KD experiments (including HARD) mainly differ in the choice of augmentation, as well as learning-rate and weight-decay. 
The plain KD experiments use Mixup ($\alpha=1.0$) and CutMix ($\alpha=1.0$) as well as AugMix \citep{hendrycks2019augmix} augmentation.
The softmax temperature was chosen as 1.0 in prior experiments and kept for all experiments.
The learning-rate for all KD and HARD experiments was chosen through a grid-search to be 0.0001 in all cases and weight-decay is 0.001 in most cases, except for HARD experiments with a ResNet18 student where a weight-decay of 0.05 is used.

\section{Knowledge Distillation Results from the Literature}
We (re-)evaluated student and teacher models from two high-performing KD experiments \citep{Beyer2021,oquab2023dinov2} in the literature on both in-domain and out-of-domain test sets.

\begin{table}[ht]
  \centering
  \caption{In-domain and out-of-domain performance for two KD experiments from the literature. Showing that the gap in out-of-domain evaluations is larger than in-domain.}
   \label{tab:literature}
  \adjustbox{width=\textwidth}{
  \begin{tabular}{lcccccccccc}
    \toprule
    \multicolumn{1}{c}{} & & \multicolumn{3}{c}{In Domain} & \multicolumn{5}{c}{Out of Domain} \\ 
    \cmidrule(lr){3-5} 
    \cmidrule(lr){6-10} 
    Model & & Val & ReaL & V2 & Im-A & Im-R & Im-C & Sketch & Style \\
    \midrule
    BiT ResNet152 (Teacher) &\multirow{ 2}{*}{\citep{Beyer2021}} &  82.9 & 87.8 & 72.0 & 31.9 & 49.2 & 51.0 & 37.4 & 16.9 \\
    BiT ResNet50 (Distilled) & & 82.8 & 87.5 & 72.5 & 25.1 & 45.3 & 51.8 & 31.6 & 15.1 \\
    \midrule
    Dino Teacher (Teacher)&\multirow{ 2}{*}{\citep{oquab2023dinov2}} &  86.5 & 89.7 & 78.4 & 76.1 & 79.3 & 27.3 & 62.8 & 34.6 \\
    ViT-S/14 (Distilled) & &  81.2 & 86.7 & 71.2 & 34.4 & 55.1 & 53.4 & 42.2 & 13.5 \\
    \bottomrule
  \end{tabular}
  }
\end{table}

\end{document}